\documentclass[letterpaper, 10 pt, conference]{ieeeconf}  
\IEEEoverridecommandlockouts                              

\usepackage{times}

\usepackage{placeins}
\usepackage{multirow}
\usepackage{makecell}
\usepackage{multicol}
\usepackage{hyperref}
\usepackage{diagbox}
\usepackage{url}
\usepackage{lipsum}
\usepackage[utf8]{inputenc} 
\usepackage[T1]{fontenc}    
\usepackage{tgtermes}
\usepackage{wrapfig}
\usepackage{tcolorbox}
\usepackage{booktabs}       
\usepackage{amsfonts}       
\usepackage{nicefrac}       
\usepackage{subcaption}
\usepackage{microtype}      
\usepackage{xcolor}
\hypersetup{
    colorlinks,
    linkcolor={blue!50!black},
    citecolor={blue!50!black},
    urlcolor={blue!80!black}
}

\usepackage{graphicx}

\usepackage{tabularray}
\usepackage{enumerate}
\usepackage{amsmath}

\usepackage{amssymb}
\usepackage{mathtools}

\usepackage{algorithm} 
\usepackage{xspace}
\usepackage{lipsum}

\usepackage{multirow}
\usepackage{makecell}
\usepackage{multicol}
\usepackage{hyperref}
\usepackage{url}
\usepackage{graphicx}
\usepackage{soul}
\usepackage{makecell}
\usepackage{caption}
\usepackage{subcaption}
\usepackage{adjustbox}
\usepackage{amsfonts}

\title{\LARGE \bf
On the Vulnerability of LLM/VLM-Controlled Robotics}

\author{Xiyang Wu$^1$, Souradip Chakraborty$^1$, Ruiqi Xian$^1$, Jing Liang$^1$, Tianrui Guan$^1$, Fuxiao Liu$^1$, \\ Brian M. Sadler$^2$, Dinesh Manocha$^1$, Amrit Singh Bedi$^3$
\thanks{$^1$ University of Maryland, College Park, MD, USA 
{\tt\small \{wuxiyang, schakra3, rxian, jingl, rayguan, fl3es, dmanocha\}@umd.edu }} %
\thanks{$^2$ DEVCOM Army Research Laboratory, Adelphi, MD, USA 
{\tt\small Brian.sadler@ieee.org }} %
\thanks{$^{3}$ University of Central Florida, Orlando, FL, USA 
{\tt\small amritbedi@ucf.edu}}
}

\usepackage{soul}
\begin{document}

\maketitle

\begin{abstract}
In this work, we highlight vulnerabilities in robotic systems integrating large language models (LLMs) and vision-language models (VLMs) due to input modality sensitivities. While LLM/VLM-controlled robots show impressive performance across various tasks, their reliability under slight input variations remains underexplored yet critical. These models are highly sensitive to instruction or perceptual input changes, which can trigger misalignment issues, leading to execution failures with severe real-world consequences.
To study this issue, we analyze the misalignment-induced vulnerabilities within LLM/VLM-controlled robotic systems and present a mathematical formulation for failure modes arising from variations in input modalities.
We propose empirical perturbation strategies to expose these vulnerabilities and validate their effectiveness through experiments on multiple robot manipulation tasks.
Our results show that simple input perturbations reduce task execution success rates by 22.2\% and 14.6\% in two representative LLM/VLM-controlled robotic systems.
These findings underscore the importance of input modality robustness and motivate further research to ensure the safe and reliable deployment of advanced LLM/VLM-controlled robotic systems.

\end{abstract}

\section{Introduction}

Large language models (LLMs) and vision-language models (VLMs) have rapidly advanced the capabilities of robotic systems, enabling robots to understand complex instructions and visual scenes. These models have shown considerable benefits across domains, from assisting in healthcare~\cite{he2023survey} and rehabilitation to optimizing manufacturing processes~\cite{wang2023chatgpt} and service tasks~\cite{felten2023will}. However, alongside these gains come substantial risks due to the inherent limitations of LLM/VLMs. For instance, language models are prone to hallucinating details~\cite{guan2023hallusionbench} or misinterpreting contextual cues~\cite{martino2023knowledge}, and when such errors occur on an embodied robot, the consequences can be serious. In this work, we highlight a surprising and critical new challenge: LLM/VLM-controlled robotic systems can be \textbf{\textit{alarmingly brittle to minor, natural variations in input modalities}}, leading to significant and unintended changes in the robot’s actions.

For example, in practical settings, a robot may receive commands from different users, each phrasing instructions in their own way. If semantically identical directives (\textit{“Pick up the red ball from the table”} vs. \textit{“Grab the red ball off the table”}) cause a robot to behave differently, it undermines reliability and could pose safety hazards. Unlike adversarial attacks – where inputs are deliberately crafted to fool the model – here, even simple, naive rephrasings by a user can inadvertently lead to a completely different outcome. This lack of robustness is not just a performance concern but a safety-critical problem, as inconsistent actions in physical environments may result in accidents or task failures.

\noindent \textbf{An Unexpected Fragility:} This instability is especially unexpected given that LLMs and VLMs are generally robust to paraphrasing and semantically similar input in other domains. A model like GPT-4, for example, will usually produce the same answer whether a user asks, \textit{“What is the capital of France?”} or \textit{“Could you tell me the capital of France?”.} We assume that meaning-preserving variations in phrasing should not drastically alter the response of a well-trained model. Indeed, in typical natural language applications, minor rewordings tend not to perturb the output significantly. 
%
%
It is, therefore, puzzling that in the context of robotics – where language models generate high-level plans for embodied agents – even minor prompt perturbations can markedly change the sequence of robot actions. This contrast suggests that integrating LLMs/VLMs with robots' task planning introduces a unique fragility absent in purely text-based tasks, stemming from their multi-modal nature. When an LLM/VLM-controlled robot fails to consistently align its understanding across modalities and language priors, \textit{e.g.} \textit{"red ball"} in the language prompt, the red ball visually perceived from the real world and the schema \textit{"red ball"} from the language prior of the embodied LLMs/VLMs on the robot, small perturbations can trigger misalignment, disrupting the entire action planning process.

\noindent \textbf{Urgency for Systematic Study:} This issue of input modality sensitivity in LLM/VLM-controlled robots represents a critical and novel challenge that warrants focused research attention. Prior work on language models in safety-critical applications has primarily centered on adversarial inputs or “jailbreak” prompts deliberately designed to trigger unwanted behaviors \cite{zou2023universal, robey2024jailbreaking}. In contrast, the failures described above stem from ordinary, well-intentioned variations in instruction or perception – a scenario that has been largely overlooked in robotics. If robots are to be trusted in homes, hospitals, and factories, they must exhibit stability and predictability regardless of how a user phrases an instruction. 
In this work, we address that gap by systematically studying and highlighting input modality sensitivity issues in state-of-the-art LLM/VLM-controlled robots. We aim to expose and analyze these concerns in depth, shedding light on how slight perturbations in the input modalities can induce failures in modern robotic systems. We seek to inform future research on building more robust and reliable LLM/VLM-controlled robots. We summarize our contributions as follows. 

\begin{figure*}[t]
    \centering
    \vspace{0.3em}
    \includegraphics[width=0.9\textwidth]{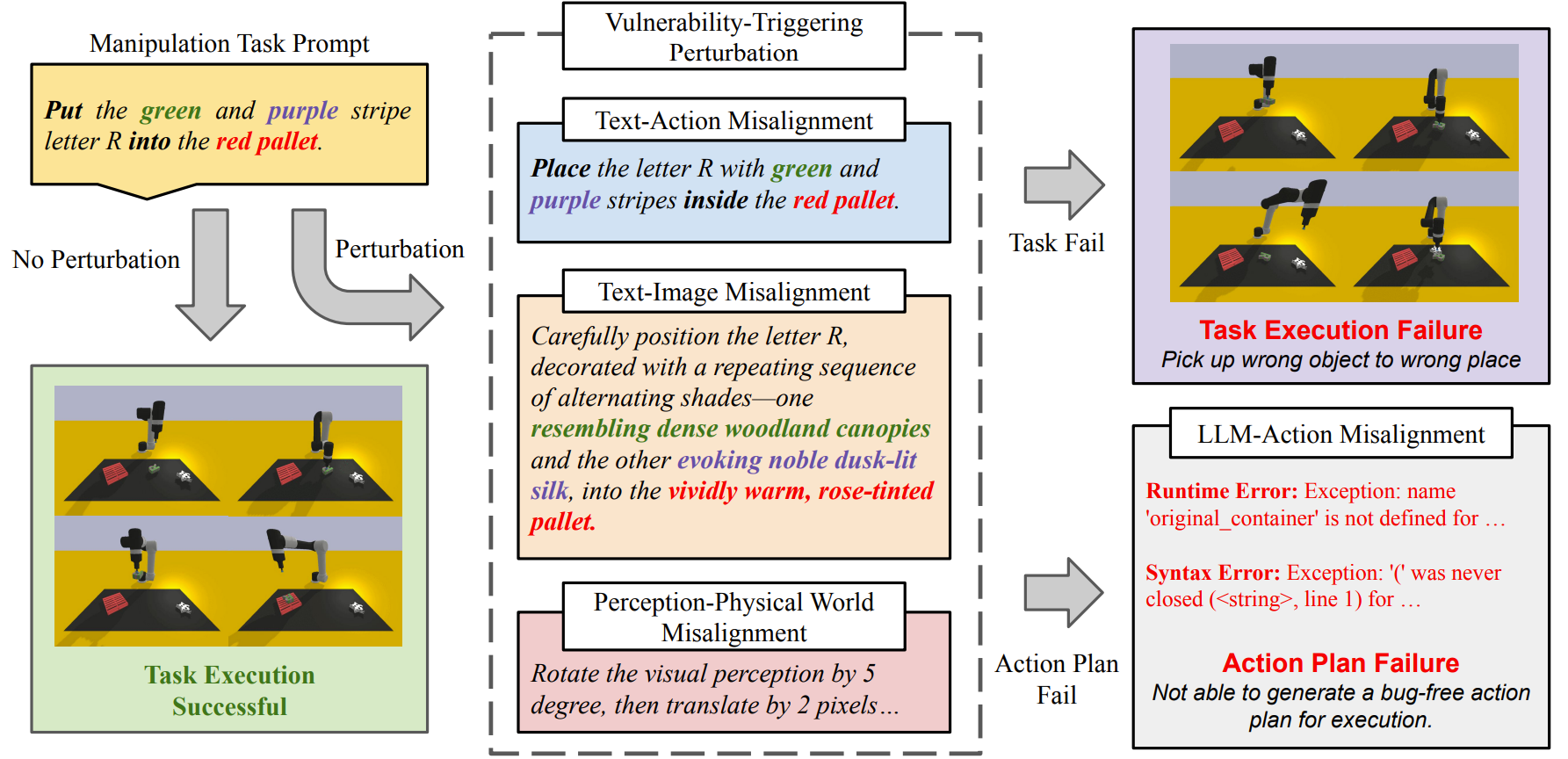}
    \caption{\textbf{Vulnerability-Triggering Perturbations.} We showcase perturbations inducing misalignment-related vulnerabilities in manipulation tasks that would otherwise succeed. These perturbations, applied to both visual and language prompt inputs, trigger misalignment-induced vulnerabilities while minimizing contextual changes:
    \textbf{(a) Text-Action Misalignment (\textcolor{blue}{blue box})} disrupts correspondence between language prompts and LLM action priors by altering action-related components with synonyms.
    \textbf{(c) Text-Image Misalignment (\textcolor{orange}{orange box})} breaks entity correspondence between prompts and visual observations by modifying entity names and attributes with synonyms or phrases.
    \textbf{(c) Perception-Physical World Misalignment (\textcolor{magenta}{magenta box})} introduces transformations to robot perceptions, misaligning them with real-world states.
    Notably, \textbf{LLM-Action misalignment} cannot be directly triggered but arise from upstream perturbations.
    Once perturbations are introduced, LLM/VLM-controlled robots are highly prone to task execution or action plan failures, significantly reducing their reliability.
    }
    \label{fig:showcase}
    \vspace{-12pt}
\end{figure*}

\noindent	\textbf{(1)	Highlighting input modality sensitivity in LLM/VLM-controlled robotic systems:} We demonstrate that current LLM/VLM-controlled robotic systems are highly sensitive to variations in input modalities. Through empirical examples, we show how minor perturbations onto the input modalities can cause dramatic changes in a robot’s behavior, sometimes triggering unsafe or undesirable actions. This reveals input modality sensitivity as a serious reliability concern in robotic applications, even without any adversarial intent.

\noindent	\textbf{(2)	Formalizing perturbation-induced failures:} 
%
We introduce a mathematical framework to characterize failures caused by input modality variations. Specifically, we define conditions where semantically similar prompts yield divergent robot behaviors. This formalization quantifies perturbation-induced instability, providing a foundation for systematically assessing an LLM/VLM-augmented robot’s sensitivity to input changes.

\noindent	\textbf{(3)	Investigating misalignment-induced vulnerabilities in state-of-the-art models:} 
We analyze the vulnerabilities in state-of-the-art LLM/VLM-controlled robotic systems that are prone to misalignment triggered by input modality variations. We propose multiple perturbation strategies to trigger these misalignment-induced vulnerabilities and validate them through experiments.
Our results show that simple perturbations in input modalities reduce success rates by 22.2\% and 14.6\% in two representative LLM/VLM-controlled robotic systems.

        

\vspace{1mm}
\noindent This work highlights the need for new methods to ensure consistent and safe robot behavior despite variations in input modalities for LLM/VLM-controlled robotic systems.

\section{Literature Review}

\subsection{Language Models for Robotics}

\noindent \textbf{Manipulation and Navigation Tasks.} The integration of LLM/VLM with robotics marks a significant advancement in embodied AI~\cite{loczson, fan2024embodied, dorbala2023can}. This fusion allows robots to leverage the commonsense and inferential capabilities of language models in decision-making tasks~\cite{song2024vlm, weerakoon2024behav, song2024tgs}. According to the criteria outlined in recent research \cite{kira2022llmroboticspaperslist, rintamaki2023everythingllmsandroboticsrepo}, the application of LLMs/VLMs in robotics primarily encompasses navigation ~\cite{ parisi2022unsurprising, huang2023visual, majumdar2020improving, payandeh2024social} and manipulation tasks~\cite{jiang2023vima, shridhar2023perceiver, bucker2023latte, brohan2023rt,liu2023mmc}. 
Recent advances in open-source vision-language-action (VLA) models for embodied robots highlight their potential in real-time decision-making. OpenVLA~\cite{kim2024openvla} is a VLA model trained on large-scale robot demonstrations, outperforming larger closed models with lower computation costs. NaVILA~\cite{cheng2024navila} integrates VLAs with locomotion skills for navigation, generating high-level commands while ensuring real-time obstacle avoidance.


 \noindent\textbf{Reasoning and Planning Tasks.} 
These tasks involve sophisticated decision-making, drawing on scene comprehension, and inherent commonsense knowledge \cite{brohan2023rt, liang2023mtg, padalkar2023open}. Enhancements in these models include pre-training for task prioritization \cite{ahn2022can} and converting complex instructions into detailed, reward-based tasks \cite{yu2023language}. These models also support human-in-the-loop decision-making, where human input refines robot demonstrations. Innovative frameworks enable robots to learn from human demonstrations and instructions \cite{shah2023mutex}, integrating large multi-modal models for better task understanding and allowing them to detect and reason over their failures once they happen~\cite{duan2024aha}.
%
%
LLM/VLM-controlled robots excel in task execution and planning but rely on well-crafted scenarios due to real-world data collection costs. 
Deploying pre-trained models for different components may cause misalignment, posing vulnerabilities in real-world deployment.


\subsection{Vulnerabilities on Language Models}
\noindent\textbf{Malfunctioning Language Models.} 
Perturbation over inputs could reliably trigger erroneous outputs from language models \cite{szegedy2013intriguing}.  \cite{liu2023covid}
involves altering model predictions through synonym replacement, random insertion, or swapping of the most influential words. 
Studies by \cite{zou2023universal, jones2023automatically} have delved into the creation of universal adversarial triggering tokens, examining their efficacy as suffixes added to input requests for language models. \cite{greshake2023youve} highlights the exploitation of language models to analyze external information, such as websites or documents, and introduces adversarial prompts through this channel. 
\cite{fu2023safety, guan2023hallusionbench, liu2023aligning} revealed vulnerabilities in language models by demonstrating the limitations of one-dimensional alignment strategies, especially when dealing with multi-modal inputs. 

\noindent\textbf{Vulnerabilities in LLM/VLM-Controlled Robots.} 
%
Substantial evidence in current literature underscores the effectiveness of LLMs/VLMs in robotics, highlighting their superior performance in various applications \cite{zhang2023large,wang2024large}.
RoboPAIR~\cite{robey2024jailbreaking} jailbreaks LLM-controlled robots, exposing safety risks in real-world deployment. 
ERT~\cite{karnik2024embodied} uses automated red-teaming to test language-conditioned robot models, revealing safety gaps. 
TrojanRobot~\cite{wang2024trojanrobot} exploits module-poisoning to backdoor vision-language robotic policies. 
\cite{wang2024ensuring} proposes a cross-layer supervision mechanism for real-time task correction and risk avoidance. 
Despite these advances, a gap remains in rigorous, mathematically grounded studies on LLM vulnerabilities in robotics. Our work addresses this by providing rigorous problem formulations, solid mathematical foundations, and empirical evidence of associated risks.

\section{Mathematical Formulation}

To study the vulnerabilities of LLM/VLM-controlled robotic systems, we also mathematically formulate the problem of the failure mode of the LLM/VLM-controlled robotic system and highlight the associated vulnerabilities.  We start by introducing the objective under which the language models are trained. For training, we follow the procedure described in \cite{jiang2023vima}, where the optimal state action trajectories are given as demonstrations denoted as $\tau = \{\tau_1, \tau_2 \cdots \tau_N\}$ where $\tau_i = \{s_0, a_0, s_1, a_1 \cdots s_T, a_T\}$ represent the $T$-length trajectory of state action pairs and the corresponding set of instructions is given by $\mathcal{I} = \{i_1, i_2 \cdots i_N\}$. Let us represent the history till the time point $t$ as $h_t = \{s_0, a_0, s_1, a_2 \cdots s_t\}$. Now, under the given setting, the optimal policy for the foundational models is obtained by maximizing the likelihood under the demonstration trajectories as
\begin{align}\label{main}
    \theta^* : = \arg \max_{\theta} \sum_{k= 1}^{N-1}\sum_{t=1}^{T-1} \log  P(a_{t}^k | s_{t}^k, h_{t}^k, i_k; \theta). 
\end{align}
In \eqref{main}, $k$ denotes the trajectory index. Once we obtain the optimal parameter $\theta^*$, our goal in this work is to study the vulnerability of the LLM/VLM-controlled robotic system under perturbations in the input modalities. Specifically, our objective is to \textbf{find vulnerability-triggering perturbations that interfere with the LLM/VLM-controlled robots to successfully accomplish task with minimal alternation in the original inputs.} To mathematically formulate that, we define the optimization problem to find out vulnerability-triggering perturbations as
\begin{align}\label{attack}
    i_{\text{perturb}} : = \arg \min_{i' \in \Omega_i}\sum_{t=1}^{T-1} \log  P(a_{t} | s_{t}, h_{t}, i'; \theta^*)
\end{align}
where, $\Omega_i$ represents the perturbation set around the original instruction $i$ given as $\Omega_i = \{i' : d(i', i) \leq \epsilon\}$ where the distance metric $d(i', i)$ ensures that the perturbed instruction $i_{\text{perturb}}$ is close under the metric $d$, which cannot be trivially filtered by a baseline defense mechanism~\cite{jain2023baseline}. This constraint restricts the instruction from being arbitrarily different, defining the validity of our perturbation of the input instruction to trigger potential vulnerabilities of LLM/VLM-controlled robots.

\noindent \textbf{Remark 1: Difference from existing LLM attacks.} We emphasize the critical difference from the standard jailbreak attacks in the context of LLMs, first introduced in \cite{zou2023universal}. In the jailbreak attacks, the target generation is fixed, which can be represented as $y^* = {y_1^*, y_2^* \cdots y_T^*}$ which can be in the context of LLMs as \textit{"Sure, this is how to make a bomb}", for the prompt $x =$ \textit{"How to make a bomb ?"}. The objective, although similar to the one defined in \eqref{attack}, has a major difference. In the case of jailbreaks, the output is fixed or targeted, and the objective is to learn $x'$ or the adversarial prompt in such a way that it has to generate the output. Thus, vanilla paraphrasing-based methods never work in the context of jailbreaks for LLMs.

On the other hand, in the case of LLM/VLM-controlled robotic systems, the perturbations causing the malfunctions of robots are inherently untargeted, and even a single change in the action can cause a significant effect on the trajectory, leading to catastrophic failure. Let us illustrate this with a simple mathematical construct as follows. Consider the trained distribution as $p_{\text{train}}$, and we assume that the probability that language model policy makes an error when the data comes from the training distribution is less than $\delta$. To formalize the notion, we assume
\begin{align}
    \textbf{prob}(a \neq \pi^*(i, h_t)) \leq \delta, \forall (i, h_t) \sim p_{\text{train}}.
\end{align}
Now, the probability of making a mistake for the trajectory of length $T$ we can characterize as
\begin{align}
    \Delta & \leq \delta T + (1-\delta)(\delta (T-1) + (1- \delta)\cdots ) \\ \nonumber
    & \approx \mathcal{O}(\delta T^2), 
\end{align}
which states that as the trajectory length for the robotic tasks increases, the probability of making mistakes with respect to changes in the input increases. For the case of out-of-distribution, the value of $\delta$ will be much higher, leading to a significant shift. This is exactly opposite to attacks on LLM/VLMs, where the purpose of the attack is to generate fixed malicious output $y*$. 

\section{Methodology}
\label{sec:problem}
\begin{figure*}[t]
    \centering
    \vspace{0.1em}
    \includegraphics[width=0.9\textwidth]{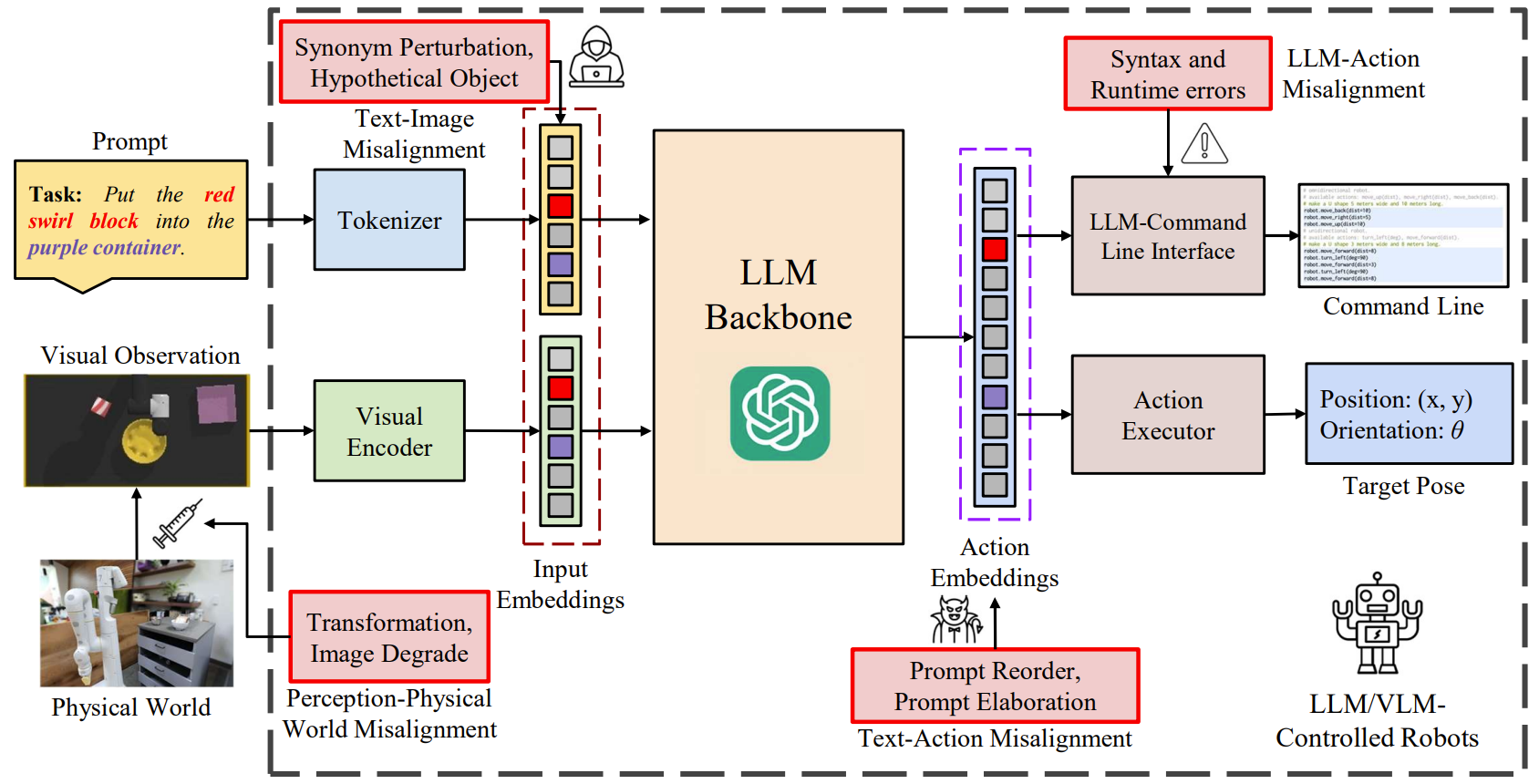}
    \caption{\textbf{Misalignment-Induced Vulnerabilities in LLM/VLM-Controlled Robots.} 
    LLM/VLM-controlled robots take language prompts and visual observations as inputs. These are processed by language tokenizers and visual encoders, mapped into the LLM’s input embedding space, while outputs are action embeddings—either command lines or target poses. Misalignments occur at four key interfaces: 
    \textbf{(a) Text-Image.} Misalignment between language and visual embeddings in the LLM input space.
    \textbf{(b) Text-Action.} Misalignment between action tokens in language prompts and the LLM’s priors.
    \textbf{(c) Perception-Physical World.} Discrepancy between the robot’s perception and real-world ground truth.
    \textbf{(d) LLM-Action.} Misalignment between the LLM’s action plans (e.g., command lines) and optimal ground-truth actions.
    }
    \label{fig:fig2}
    \vspace{-10pt}
\end{figure*}

\subsection{A Deep Dive into LLM/VLM-Controlled Robots}

In this section, we first examine trends in LLM/VLM-controlled robot architectures before highlighting key vulnerabilities. LLM/VLM-controlled robotic systems, often termed vision-language-action (VLA) models, belong to the multi-modal foundation model family~\cite{ma2024survey, li2025benchmark}. Like LLaVA~\cite{liu2023visual} and Flamingo~\cite{alayrac2022flamingo}, they incorporate the following components (Figure \ref{fig:fig2}):

\begin{itemize} 
    \item \textbf{Vision Encoder.} Converts image-based observations into embeddings, typically using an adapter network. Object segmentation is often included for scene understanding. 
    \item \textbf{Language Tokenizer.} Translates natural language prompts into the backbone LLM’s input domain. 
    \item \textbf{Backbone LLM.} Processes multi-modal inputs and generates executable action plans, producing either goal poses~\cite{brohan2023rt, jiang2023vima, kim2024openvla} or code-based commands~\cite{huang2023instruct2act, yu2023language, liang2023code}. 
    \item \textbf{Action Executor.} A predefined policy executes the generated action plan. \end{itemize}


An ideal LLM/VLM-controlled robot, trained on vast real-world interactions, flawlessly understands input contexts and executes optimal plans. Statistically, the distributions of input modalities are aligned, while the distribution of the output action plans is aligned with optimal decisions for given tasks. Additionally, each component's input distribution should match its upstream output. \textbf{Alignment is crucial for the LLM/VLM-controlled robotic system’s performance.}

\subsection{Misalignment-Induced Vulnerabilities}
\label{sec:methodology:alignment}
However, due to limited high-quality robotics datasets and costly model training, most works incorporate pre-trained models as components in LLM/VLM-controlled robotic systems, such as open-source vision encoders (e.g., CLIP~\cite{zhou2024aligning}, ViT~\cite{dosovitskiy2020image}) and LLM backbones like GPT~\cite{achiam2023gpt} or LLAMA~\cite{dubey2024llama}.
Pre-trained models drive advancements in LLM/VLM-controlled robots but also introduce vulnerabilities in control tasks. Gaps in the training datasets of vision encoders, tokenizers, and LLMs~\cite{zhou2024aligning} can make these models highly sensitive to slight input perturbations, triggering misalignment issues which further cause the failure of robot task execution. Key sources of vulnerabilities include:
\begin{itemize}
    \item \textbf{Text-Image Misalignment.} These vulnerabilities arise when the LLM fails to associate entities in language prompts with those in visual observations. 
    For example, \textit{vibrant, crimson block adorned with mesmerizing swirling patterns} and \textit{red swirl block} should be treated as synonyms, with their embeddings in the LLM’s input space aligned, alongside the visually perceived \textit{red swirl block} entity.
    However, misaligned LVLMs may interpret them as distinct objects.
    \item \textbf{Text-Action Misalignment.} LLM/VLM-controlled robots often rely on rigidly structured instructions in describing actions (\textit{e.g.}, \textit{Put \{Object A\} to \{Position B\}}). Even minor paraphrasing (\textit{e.g.}, \textit{Place \{Object A\} inside \{Position B\}}) can lead to severe misinterpretations by introducing misalignment between the language prompt space and the executable action space.
    \item \textbf{Perception-Physical World Misalignment.} Due to the Sim2Real gap, as robots are trained with collected datasets or crafted simulators rather than actual interactions with the real world, robots perceive environments differently from humans. Statistically, the robot's perception distribution may be misaligned with the actual state space distribution in the real world. A robot retrieving a red block relies on coordinates \textit{(x, y)}, assuming it remains there. If moved to \textit{(x', y')}, the robot may fail, whereas humans locate the object based on perception rather than fixed coordinates. Similar challenges arise in scene understanding and captioning.  
    \item \textbf{LLM-Action Misalignment.} Command-line action executors introduce risks due to LLM misinterpretations. Unlike standard code generation with LLM~\cite{jiang2024survey}, code generation for robotic execution requires a precise understanding of functional tools and objects. Misalignments between LLM and the command-line action executors, stemming from sparse training samples and generic LLM frameworks not tailored for robot-specific tasks, can lead to syntax errors (\textit{e.g.} incorrect variable names, compilation failures) and runtime errors (\textit{e.g.} function misunderstandings, failure to map perceived objects across modalities). 
\end{itemize}

\subsection{Vulnerability-Triggering Perturbations}
\label{sec:methodology:perturbation}
Targeting the misalignment-induced vulnerabilities in Section~\ref{sec:methodology:alignment}, we design vulnerability-triggering perturbations to induce robot failures during task execution. Treating the robotic system as a black box, we focus on perturbing input modalities of LLM/VLM-controlled robots, without modifying model parameters or intermediate results. Our perturbation strategies include:

\begin{itemize}
    \item \textbf{P1:} Perturbations triggering the text-image misalignment pattern focus on breaking the correspondence between entities in language and visual modalities. We replace essential components describing the objects within the input prompt with their synonyms, targeting entity names and attributes. 
    \item \textbf{P2:} Perturbations targeting the text-action alignment intend to distract the action understanding of the LLM backbone. We modify action-related components inside the input prompt by synonym replacement, reordering, or adding excessive descriptive details without altering task intent.  
    \item \textbf{P3:} Perturbations for perception-physical world misalignment focus on interfering the pre-defined, highly artificial correspondence between robot perception and the physical world, while our perturbations ensure robots accomplish the task if they stick to their pre-perturbation action plans. One perturbation strategy is to induce slight, undetectable shifts in object positions rather than significantly changing the layout of the perception. 
    \item \textbf{P4:} Since LLM-based code generation is a black box, our perturbation strategies on input modalities cannot directly deploy on the code generation. However, we conduct experiments comparing two action execution strategies (goal-reaching vs. command-line) on the same benchmark to reveal how input perturbations propagate to action execution and trigger LLM-action misalignments.  
\end{itemize}

\section{Experimental Evidence}
\subsection{Experiment Overview}

We investigate vulnerabilities in LLM/VLM-controlled robotic systems caused by misalignment and identify perturbations that trigger these vulnerabilities. Our experiments focus on two representative systems for manipulation tasks: VIMA~\cite{jiang2023vima}, which employs a goal-reaching action planner, and Instruct2Act~\cite{huang2023instruct2act}, which generates executable command lines as action plans. Our objectives include:

\begin{itemize}
    \item Assess the severity of misalignment-induced vulnerabilities using our proposed perturbations across tasks that vary in reliance on perception and reasoning.
    \item Evaluate the robustness of LLM/VLM-controlled robotic systems under perturbations across manipulation tasks with different levels of generalization, reasoning, planning, and context understanding.
\end{itemize}

\noindent \textbf{Perturbations.}
Here we provide the specific details of perturbations we introduce for experiments. Ideally, a well-aligned system should execute tasks flawlessly despite these perturbations:

\begin{itemize}
    \item \textbf{P1 for Text-Image Misalignment. } (1) Entity Perturbation (Entity): Replacing entities in prompts with synonyms. (2) Attribute Perturbation (Attribute): Substituting descriptive attributes with synonyms. (3) Hypothetical Object Insertion (Hypo. Obj.): Adding a non-task-related object to perception to test scene understanding.
    \item \textbf{P2 for Text-Action Misalignment.} (1) Reorder the Prompt (Reorder): Paraphrasing and altering action-related words. (2) Elaborate the Prompt (Elaborate): Adding excessive descriptive details.
    \item \textbf{P3 for Perception-Physical World Misalignment.} We investigate two perturbations for Text-action Misalignment: (1) Transform the Perception (Transform): Applying slight image transformations to shift perceived object positions: (2) Degrade the Perception (Degrade): Lowering perception quality to distort object recognition.
    \item \textbf{P4 for LLM-Action Misalignment.} We perturb input modalities to examine scene understanding, object correspondence, and action planning, inducing misalignments between LLM outputs and actual actions. Comparing goal-reaching and command-line action planning under the same benchmark, we analyze how perturbations propagate to execution failures.
\end{itemize}

\begin{table*}[t]
  \vspace{1em}
  \begin{center}
  \resizebox{\textwidth}{!}{
    \begin{tabular}{cccccccccccccccccccccccccc}
    \toprule
        \multicolumn{2}{c}{} & \multicolumn{4}{c}{\makecell{Visual  Manipulation}} & \multicolumn{4}{c}{\makecell{Scene Understanding}}  & \multicolumn{4}{c}{\makecell{Sweep w/o. Exceeding}} & \multicolumn{4}{c}{\makecell{Pick in order then Restore}}  \\
        \cmidrule(lr){3-6} \cmidrule(lr){7-10} \cmidrule(lr){11-14} \cmidrule(lr){15-18}
         \makecell{Misalignment}  & \makecell{Perturbation}  & \makecell{Input \\ Sim.}& \makecell{Action \\ CosSim.} & \makecell{VIMA \\ SR} & \makecell{I2A \\ SR} & \makecell{Input \\ Sim.} & \makecell{Action \\ CosSim.} & \makecell{VIMA \\ SR} & \makecell{I2A \\ SR} & \makecell{Input \\ Sim.} & \makecell{Action \\ CosSim.} & \makecell{VIMA \\ SR} & \makecell{I2A \\ SR} & \makecell{Input \\ Sim.} & \makecell{Action \\ CosSim.} & \makecell{VIMA \\ SR} & \makecell{I2A \\ SR} \\
        \midrule
        \multirow{3}*{\makecell{Text-Image}} & \makecell{Entity} & 0.993 & 0.760 & 66.7 & 26.2 & 1.000 & 0.931 & 90.7 & 8.6 & 1.000 & 0.944 & 90.0 & 10.3 & 1.000 & 0.868 & 8.7 & 0.0 \\
        ~ & \makecell{Attribute} & 0.987 & 0.786 & 66.7 & 43.3 & 1.000 & 0.948 & 94.0 & 10.1 & 0.966 & 0.950 & 88.7 & 0.0 & 0.993 & 0.850 & 10.7 & 0.0 \\
        ~ &\makecell{Hypo. Obj.} & 0.974 & 0.887 & 82.4 & 41.1 & 0.975 & 0.836 & 88.4 & 30.9 & 0.974 & 0.928 & 87.4 & 13.8 & 0.976 & 0.967 & 25.7 & 0.0 \\
        \midrule
        \multirow{2}*{\makecell{Text-Action}} & \makecell{Reorder} & 1.000 & 0.832 & 76.7 & 23.9 & 1.000 & 0.992 & 100.0 & 20.6 & 0.993 & 0.945 & 88.7 & 20.7 & 1.000 & 0.860 & 16.0 & 0.0 \\
        ~ & \makecell{Elaborate} & 1.000 & 0.792 & 66.0 & 21.1 & 1.000 & 0.958 & 95.3 & 12.0 & 0.993 & 0.937 & 88.7 & 6.9 & 0.993 & 0.859 & 8.7 & 0.0 \\
        \midrule
        \multirow{2}*{\makecell{Perception- \\ Physical World}} &  \makecell{Transform} & 0.844 & 0.445 & 33.0 & 16.4 & 0.822 & 0.367 & 29.5 & 13.0 & 0.853 & 0.465 & 58.5 & 16.4 & 0.678 & 0.726 & 5.0 & 1.7 \\
        ~ & \makecell{Img. Degrade} & 0.560 & 0.976 & 97.8 & 12.1 & 0.563 &	0.973 & 100.0 & 10.0 & 0.572 & 0.967 & 92.9 & 18.4 & 0.482 & 0.959 & 26.6 & 1.2 \\
        \midrule
        \multicolumn{2}{c}{Origin} & - & - & 98.7 & 47.4 & - & - & 100.0 & 39.6 & - & - & 94.7 & 20.7 & - & - & 48.0 & 3.4 \\
    \bottomrule
    \end{tabular}
    }
    \end{center}
    \caption{\textbf{Vulnerability-Triggering Perturbations.} We perform evaluation experiments under vulnerability-triggering perturbations targeting on both VIMA~\cite{jiang2023vima} and Instruct2Act (I2A)~\cite{huang2023instruct2act} over $4$ tasks on VIMA-Bench: \textit{Visual Manipulation}, \textit{Scene Understanding}, \textit{Sweep without Exceeding}, and \textit{Pick in order then Restore}. \textbf{Conclusion: } 
    Both LLM/VLM-controlled robotic systems are vulnerable to \textit{Perception-Physical World} misalignments. VIMA demonstrates greater robustness in context-understanding tasks such as \textit{Scene Understanding} and \textit{Sweep without Exceeding}, whereas Instruct2Act excels in planning-heavy tasks like \textit{Pick in order then Restore}.
    }
    \label{tab:results}
\end{table*}

\begin{table*}[t]
  \begin{center}
  \resizebox{\textwidth}{!}{
    \begin{tabular}{cccccccccccccccccccccccccc}
    \toprule
        ~ & ~ & \multicolumn{8}{c}{\makecell{Combinatorial Generalization}} & \multicolumn{8}{c}{\makecell{Novel Object Generalization}} \\
        \cmidrule(lr){3-10} \cmidrule(lr){11-18}
        ~ & ~ & \multicolumn{4}{c}{\makecell{Visual  Manipulation}} & \multicolumn{4}{c}{\makecell{Pick in order then Restore}}  & \multicolumn{4}{c}{\makecell{Visual  Manipulation}}
        & \multicolumn{4}{c}{\makecell{Pick in order then Restore}}  \\
        \cmidrule(lr){3-6} \cmidrule(lr){7-10} \cmidrule(lr){11-14} \cmidrule(lr){15-18}
         \makecell{Misalignment}  & \makecell{Perturbation}  & \makecell{Input \\ Sim.}& \makecell{Action \\ CosSim.} & \makecell{VIMA \\ SR} & \makecell{I2A \\ SR} & \makecell{Input \\ Sim.} & \makecell{Action \\ CosSim.} & \makecell{VIMA \\ SR} & \makecell{I2A \\ SR} & \makecell{Input \\ Sim.} & \makecell{Action \\ CosSim.} & \makecell{VIMA \\ SR} & \makecell{I2A \\ SR} & \makecell{Input \\ Sim.} & \makecell{Action \\ CosSim.} & \makecell{VIMA \\ SR} & \makecell{I2A \\ SR} \\
        \midrule
        \multirow{3}*{\makecell{Text-Image}} & \makecell{Entity} & 1.000 & 0.773 & 62.0 & 25.0 & 1.000 & 0.865 & 8.0 & 4.2 & 1.000 & 0.703 & 48.0 & 58.3 & 1.000 & 0.854 & 0.0 & 8.3 \\
        ~ & \makecell{Attribute} & 0.980 & 0.771 & 62.7 & 37.5 & 1.000 & 0.854 & 7.3 & 0.0 & 0.980 & 0.693 & 44.0 & 33.3 & 0.987 & 0.854 & 0.0 & 4.2 \\
        ~ &\makecell{Hypo. Obj.} & 0.974 & 0.890 & 81.4 & 55.2 & 0.977 & 0.970 & 23.0 & 20.7 & 0.974 & 0.890 & 81.4 & 66.7 & 0.961 & 0.962 & 2.3 & 54.2 \\
        \midrule
        \multirow{2}*{\makecell{Text-Action}} & \makecell{Reorder} & 0.993 & 0.868 & 74.6 &  51.7 & 1.000 & 0.857 & 12.0 & 24.1 & 0.993 & 0.745 & 59.3 & 75.9 & 1.000 & 0.817 & 0.0 & 38.0 \\
        ~ & \makecell{Elaborate} & 1.000 & 0.788 & 62.7 & 55.2 & 1.000 & 0.856 & 9.3 & 13.8 & 0.993 & 0.704 & 46.0 & 62.1 & 0.993 & 0.827 & 0.0 & 34.5 \\
        \midrule
        \multirow{2}*{\makecell{Perception- \\ Physical World}} &  \makecell{Transform} & 0.839 & 0.455 & 32.7 & 56.9 & 0.672 & 0.731 & 3.7 & 24.1 & 0.828 & 0.463 & 29.8 & 69.8 & 0.609 & 0.734 & 4.3 & 47.9 \\
        ~ & \makecell{Img. Degrade} & 0.560 & 0.977 & 96.0 & 54.0 & 0.574 & 0.974 & 20.4 & 21.8 & 0.562 & 0.985 & 91.3 & 68.1 & 0.564 & 0.970 & 0.9 & 51.4 \\
        \midrule
        \multicolumn{2}{c}{Origin} & - & - & 96.7 & 58.6 & - & - & 39.3 & 31.0 & - & - & 95.0 & 79.3 & - & - & 6.0 & 54.2 \\
    \bottomrule
    \end{tabular}
    }
    \end{center}
    \caption{\textbf{Vulnerability-Triggering Perturbations under Different Generalization Levels.} We perform evaluation experiments under vulnerability-triggering perturbations targeting on both VIMA~\cite{jiang2023vima} and Instruct2Act (I2A)~\cite{huang2023instruct2act} over $2$ tasks on VIMA-Bench: \textit{Visual Manipulation}, and \textit{Pick in order then Restore} over $2$ higher generalization levels \textit{Combinatorial Generalization} and \textit{Novel Object Generalization}, apart from the \textit{Placement Generalization} included in Table~\ref{tab:results}. \textbf{Conclusion:} LLM/VLM-controlled robotic systems using the command-line action execution policy are more robust under perturbation when task and scene complexity increases.}
    \label{tab:results_general}
    \vspace{-15pt}
\end{table*}

\noindent \textbf{Benchmarks.} 
We conduct extensive experiments on various robot manipulation tasks to evaluate different aspects of LLM/VLM-controlled systems. Using VIMA-Bench~\cite{jiang2023vima}, we test four tasks: \textit{Visual Manipulation}, \textit{Scene Understanding}, \textit{Sweep without Exceeding}, and \textit{Pick in order then Restore}, assessing LLM/VLM-controlled robotic systems' abilities over visual reasoning, scene understanding, and action planning.
Experiments cover three generalization levels: \textit{Placement Generalization}, \textit{Combinatorial Generalization}, and \textit{Novel Object Generalization}, ranked by the complexity of manipulation tasks encountered and the contextual information contained in interactions with entities involved~\cite{jiang2023vima}. Our goal is to analyze how misalignment-induced vulnerabilities vary across tasks, as each relies on different system capabilities, making LLM/VLM-controlled robots susceptible in distinct ways.


\noindent \textbf{Evaluation Metrics.} 
To assess our vulnerability-triggering perturbations, we use three key metrics: input similarity, action embedding similarity, and task success rate.\\
\noindent \textbf{(a) Input Similarity (Input Sim.)} measures contextual distance before and after perturbation. GPT-4-Turbo~\cite{yang2023dawn} evaluates prompt consistency, while SSIM assesses visual similarity.\\
\noindent \textbf{(b) Action Cosine Similarity (Action CosSim.)} is computed via cosine similarity, aiming to maximize action embedding differences post-perturbation.\\
\noindent \textbf{(c) Task Success Rate (SR)}, measured over 150 tasks, evaluates system robustness under perturbations.


\subsection{Results over Vulnerability-Triggering Perturbations}
\label{sec:results}

Tables~\ref{tab:results} and~\ref{tab:results_general} present experimental results on multiple vulnerability-triggering perturbations across four manipulation tasks and three generalization levels, focusing on context perception, comprehension, and reasoning in LLM/VLM-controlled robots. While most input modality similarity scores are high, indicating minimal contextual variation before and after perturbation, success rates for both models vary significantly across tasks. Our analysis provides key insights into these vulnerabilities:

\noindent \textbf{1. Vulnerability from Perception-Physical World Misalignment.} 
Among all misalignments causing task failures presented in Table~\ref{tab:results}, LLM/VLM-controlled robots are most vulnerable to perception-physical world misalignments. 
%
VIMA's success rate drops by 29.9\% on average, while Instruct2Act sees a sharp 21.5\% drop across four tasks. In contrast, text-image and text-action misalignments cause milder drops of 18.7\% and 17.8\% for VIMA, while Instruct2Act shows a similar trend. This suggests greater robustness to language prompt perturbations than visual perception changes.
%
This phenomenon stems from the Sim2Real gap, arising from how LLM/VLM-controlled robots perceive the physical world. As discussed in Section~\ref{sec:methodology:alignment}, these robots struggle to interpret slight perception variations, and even minor deviations can disrupt action planning by significantly altering the robot’s understanding of the environment.
On the other hand, even though LLM/VLM-controlled robots may not encounter a sufficient number of entities or action variations in their training, their pre-trained LLM backbone models retain some semantic understanding from their language priors, helping mitigate misalignments. This is reflected in better success rates under perturbations targeting text-image and text-action misalignments.



\noindent \textbf{2. VIMA's Robustness in Context Understanding Tasks.} 
%
Among the tested manipulation tasks, \textit{Scene Understanding} and \textit{Sweep without Exceeding} require strong scene understanding, such as aligning contextual references between language prompts and visual perceptions (\textit{Scene Understanding}) or understanding constraints based on spatial relationships in the physical world (\textit{Sweep without Exceeding}).
As shown in Table~\ref{tab:results}, VIMA demonstrates strong robustness to text-action and text-image misalignments, with a success rate drop of less than 9\% on both tasks. In contrast, Instruct2Act, which relies on command-line planning, suffers a substantial 23\% drop.
Both tasks emphasize visual understanding abilities for LLM/VLM-controlled robots, where VIMA excels, likely due to its greater exposure to vision-dependent tasks in its training data that help mitigate misalignments. 
On the other hand, Instruct2Act, which directly employs GPT-4-Turbo for off-the-shelf command-line-based action planning, is more reliant on language-based training data. This increases the risk of cross-modal misalignment, further degrading its performance.

\noindent \textbf{3. Models' Vulnerability in Planning-Heavy Tasks.} 
%
\textit{Pick in order then Restore} is a planning-intensive task requiring multi-step execution. Complexity increases with new entities and tasks, demanding higher generalization. As shown in Table~\ref{tab:results} and~\ref{tab:results_general}, VIMA's success rate drops from 48.0\% to 6.0\%, while Instruct2Act improves from 3.4\% to 54.2\%. Perturbations further reduce VIMA’s success rate by 30.5\%, whereas Instruct2Act experiences a smaller 16.3\% drop on average.
%
This discrepancy stems from the backbone LLM's reasoning and planning abilities, which scales with its parameter count. As task complexity rises, LLM/VLM-controlled robots depend more on the backbone LLM for decision-making rather than visual perception or language prompts. 
This benefits Instruct2Act, whose GPT-4-Turbo-powered command-line action planner produces reliable plans. In contrast, VIMA, despite incorporating historical actions in planning, struggles with hierarchical action generation.

\noindent \textbf{4. The Impact of Generalization on Models' Vulnerabilities.}
%
%
Higher generalization levels introduce novel entities and tasks, increasing complexity and demanding stronger context understanding and reasoning. As shown in Table~\ref{tab:results_general}, VIMA experiences a significant 25.0\% drop in success rate across all perturbations, highlighting its severe vulnerability to misalignments. In contrast, Instruct2Act shows a smaller 14.3\% drop, demonstrating superior robustness due to its stronger LLM backbone.
Breaking down vulnerabilities by misalignment type, Instruct2Act remains highly resistant to perception-physical world misalignments (only a 6.5\% drop) but is notably weaker against text-image misalignments (25.1\% drop). This phenomenon stems from its command-line action planner, which relies on entity alignments. Perturbations disrupting entity alignment can significantly degrade its performance. However, Instruct2Act's reliance on contextual entity information rather than spatial positioning enhances robustness against perception-physical world misalignments, making it less sensitive to visual perception distortions.

\subsection{Discussion}
\label{sec:result_discusion}
Our comprehensive experiments and analysis provide deeper insights into vulnerabilities induced by perturbations on the input modalitie and potential improvements for LLM/VLM-controlled robotic systems. Our keynotes include:

\noindent \textbf{1. Task-Specific Vulnerabilities.} 
Different tasks emphasize distinct aspects of LLM/VLM-controlled robotic systems, as varying module contributions influence decision-making. Consequently, the impact of vulnerability-triggering perturbations differs across tasks. Overall, Instruct2Act exhibits greater robustness in planning-heavy tasks, while VIMA is more reliable in context-understanding tasks.

\noindent \textbf{2. The Role of Pre-Training.} 
Our experiments compare VIMA, which uses a small-scale backbone model trained on manipulation tasks within the benchmark distribution, and Instruct2Act, which leverages an off-the-shelf general-purpose LLM. Results indicate that VIMA, benefiting from domain-aligned training data, excels in context understanding, while Instruct2Act outperforms in action planning and reasoning. This gap stems from differences in their pre-training datasets.

\noindent \textbf{3. Need for Improved Alignment and Data Sufficiency.} 
%
Our results highlight the need for further investigation to ensure consistent and safe robot behavior despite variations, where key challenges include improving cross-modality alignment in LLM/VLM-controlled robots and addressing the scarcity of diverse, high-quality training data. Current vulnerabilities stem from misalignments due to limited data coverage across potential scenarios. Strengthening cross-modal alignment, expanding datasets, and leveraging high-fidelity synthetic training are essential to mitigate vulnerabilities induced by input modality perturbations.

\section{Conclusions}

In this study, we investigate vulnerabilities in LLM/VLM-controlled robots, where small input perturbations can lead to severe task failures. We rigorously formulate the problem in searching for vulnerability-triggering perturbations with a solid mathematical foundation, based on our analysis of the structures for the LLM/VLM-controlled robotic systems and misalignment-induced vulnerabilities across modalities and language priors. 
We propose multiple perturbation strategies to trigger these vulnerabilities within LLM/VLM-controlled robotic systems, and we validate their effectiveness
by conducting experiments on multiple robot manipulation tasks. Our results show that LLM/VLM-controlled robots are highly sensitive to crafted perturbations, with vulnerabilities varying by task and model. Our future work will further explore misalignment-induced vulnerabilities in LLM/VLM-controlled robotic systems, develop automated vulnerability-triggering mechanisms, and integrate them into model training to enhance the robustness of future LLM/VLM-controlled robotic systems.


\footnotesize{
\bibliographystyle{ieeetr}
\bibliography{iclr2024_conference}
}


\end{document}